\documentclass[letterpaper,10pt,conference]{ieeeconf}

\usepackage{copyright}

\IEEEoverridecommandlockouts

\usepackage{pifont}

\usepackage{graphicx}
\usepackage{booktabs}

\usepackage{amsmath}
\usepackage{amssymb}
\usepackage{mathtools}
\usepackage{physics}
\usepackage{xfrac}

\let\labelindent\relax
\usepackage[inline]{enumitem}

\usepackage{pifont}
\newcommand{\xmark}{\ding{55}}%

\usepackage{threeparttable}
\usepackage{tabularx}

\usepackage[labelformat=simple]{subcaption}
\DeclareCaptionLabelSeparator{periodspace}{.\quad}
\usepackage{multirow}
\usepackage{lipsum}  
\usepackage{duckuments}

\DeclarePairedDelimiterX\set[1]\lbrace\rbrace{\def\given{\;\delimsize\vert\;}#1}

\usepackage{cleveref}
\crefname{table}{Tab.}{Tabs.}
\crefname{figure}{Fig.}{Figs.}
\crefname{section}{Sec.}{Secs.}
\crefname{equation}{Eq.}{Eqs.}

\usepackage[acronym]{glossaries}
\newacronym{auc}{AUC}{area-under-curve}
\newacronym{ad}{AD}{Autonomous Driving}
\newacronym[longplural=Degrees-of-Freedom]{dof}{DoF}{Degree-of-Freedom}
\newacronym{svd}{SVD}{Singular Value Decomposition}
\newacronym{ransac}{RANSAC}{RANdom SAmple Consensus}
\newacronym{dbscan}{DBSCAN}{Density-Based Spatial Clustering of Applications with Noise}
\newacronym{mlp}{MLP}{Multi-Layer Perceptron}
\newacronym{rmse}{RMSE}{Root Mean Square Error}
\newacronym{icp}{ICP}{Iterative Closest Point}
\newacronym{fpfh}{FPFH}{Fast Point Feature Histograms}
\newacronym{fcgf}{FCGF}{Fully Convolutional Geometric Features}
\newacronym{dgr}{DGR}{Deep Global Registration}
\newacronym{cnn}{CNN}{Convolutional Neural Network}
\newacronym{miou}{mIoU}{mean Intersection over Union}
\newacronym{rte}{RTE}{Relative Translation Error}
\newacronym{rre}{RRE}{Relative Rotation Error}
\newacronym{gicp}{GICP}{Generalised Iterative Closest Point}
\newacronym{iss}{ISS}{Intrinsic Shape Signatures}
\newacronym{kpg}{KPG}{KeyPoint Quality}
\newacronym{dcp}{DCP}{Deep Closest Point}
\newacronym{dcpcr}{DCPCR}{Deep Compressed Point Cloud Registration}
\newacronym{mdgat}{MDGAT}{Multiplex Dynamic Graph ATtention network}
\newacronym{gnn}{GNN}{Graph Neural Network}

\usepackage{cuted}

\usepackage{tikz}
\usetikzlibrary{fit,calc}

\usepackage{float}

\usepackage{pgf} 
\usepackage{pgfplots}
\usepgfplotslibrary{external}
\usepackage{pgfplotstable}
\usepackage{pgfmath}
\usepgfplotslibrary{statistics}
\pgfplotsset{compat=1.8}

\usepackage{filecontents}

\usepackage[binary-units]{siunitx}

\usepackage{scalefnt}

\usepackage{xcolor}

\begin{document}

\title{
\Large
\bf
That’s My Point: Compact Object-centric LiDAR Pose Estimation \\for Large-scale Outdoor Localisation
}
\author{Georgi Pramatarov, Matthew Gadd, Paul Newman, and Daniele De Martini
\\
Mobile Robotics Group (MRG), University of Oxford\\\texttt{\{georgi,mattgadd,pnewman,daniele\}@robots.ox.ac.uk}
\thanks{This work was supported by EPSRC Programme Grant ``From Sensing to Collaboration'' (EP/V000748/1).
}}
\maketitle

\copyrightnotice

\begin{abstract}

This paper is about 3D pose estimation on LiDAR scans with extremely minimal storage requirements to enable scalable mapping and localisation.
We achieve this by clustering all points of segmented scans into semantic objects and representing them only with their respective centroid and semantic class.
In this way, each LiDAR scan is reduced to a compact collection of four-number vectors.
This abstracts away important structural information from the scenes, which is crucial for traditional registration approaches.
To mitigate this, we introduce an object-matching network based on self- and cross-correlation that captures geometric and semantic relationships between entities.
The respective matches allow us to recover the relative transformation between scans through weighted \gls{svd} and \gls{ransac}.
We demonstrate that such representation is sufficient for metric localisation by registering point clouds taken under different viewpoints on the KITTI dataset, and at different periods of time localising between KITTI and KITTI-360.
We achieve accurate metric estimates comparable with state-of-the-art methods with almost half the representation size, specifically \SI{1.33}{\kilo\byte} on average.

\end{abstract}

\glsresetall

\section{Introduction}%
\label{sec:introduction}

Localisation is crucial for mobile robotics, allowing safe autonomous motion.
For this task, LiDAR is popular due to its intrinsically geometric readings and robustness to lighting and appearance, thus providing a very informative and stable representation of its surroundings.

A common approach to localisation is creating a map from a previous traversal of the environment and then registering live sensor readings onto the map to extract the metric ego-vehicle pose.
However, modern LiDAR sensors can collect up to tens of thousands of points per scan, making the representation and storage of \textit{compressed but reliable} LiDAR observations compelling.
Especially in the \gls{ad} domain, maps are prone to being vast and may require a substantial amount of memory to scale~\cite{huang2020octsqueeze}.
This is also critical in distributed settings, where the map and observations need to be repeatedly transmitted between multiple agents and/or servers~\cite{cieslewski2018dataefficient, ramtoula2020capricorn}.

\begin{figure}
\centering
\includegraphics[width=0.925\columnwidth]{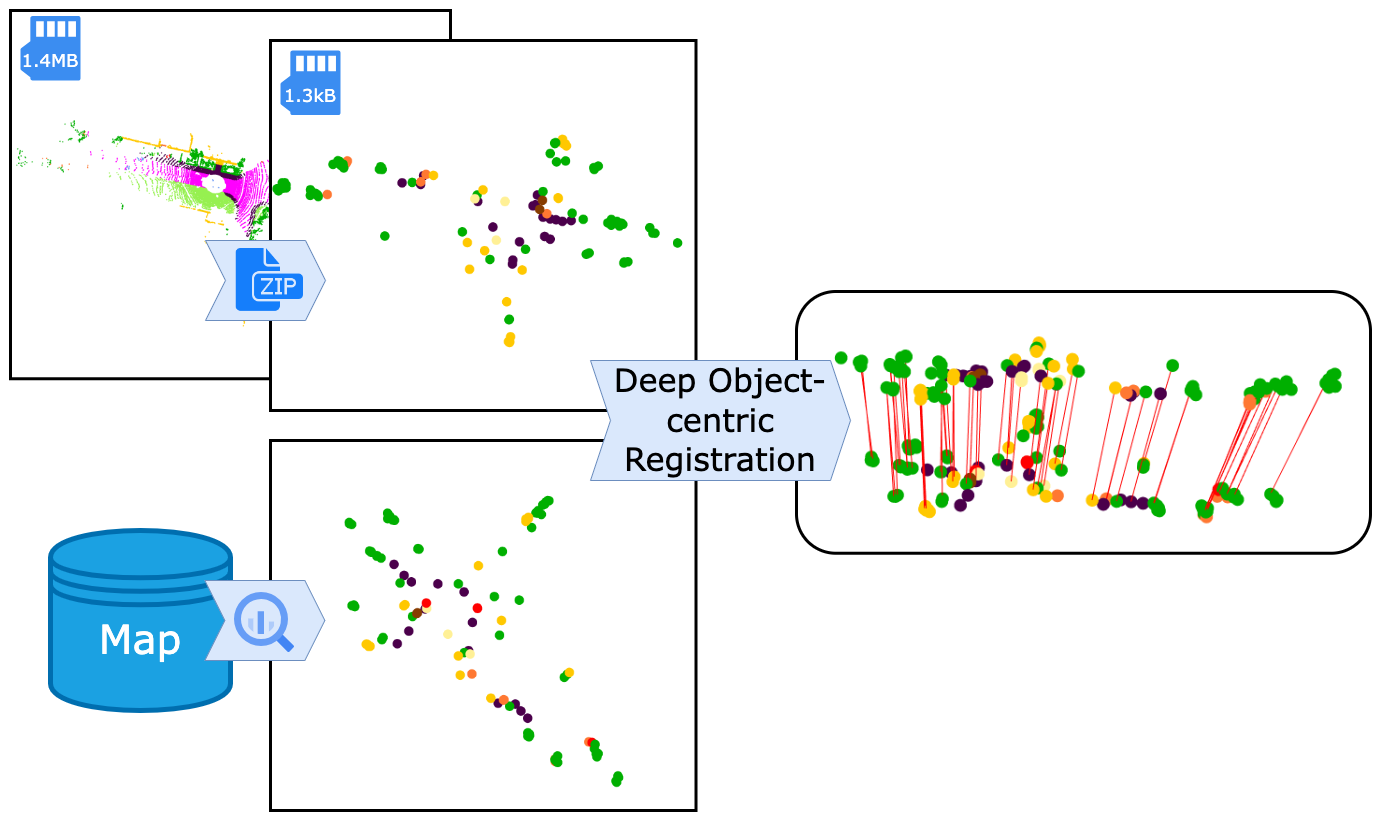}
\caption{
Method overview.
LiDAR scans are represented extremely compactly by only the centroid and semantic class of the corresponding objects in the scene.
Sacrificing information in this way, we learn a robust matching function which leverages the remaining geometry in the object scene structure as well as the semantic relationships between entities. 
}
\label{fig:overview}
\vspace{-1cm}
\end{figure}

Whilst some methods use pure geometric approaches \cite{huang2020octsqueeze,cao20193d}, semantics and objects \cite{pramatarov2022boxgraph} provide a tradeoff between emphasising local keypoints or global features -- while at the same time being human understandable \cite{panagiotaki2023sem}.
Each object can be described numerically, and scan matching and registration rely on accurate object correspondences.
Motivated by these approaches, this work tries to answer the question: how small can an object descriptor be while still providing enough information for a reliable and accurate scan registration?
Here, we show that a descriptor composed of the positions of the objects' centroids and their class -- i.e. three floating-points and one byte -- is enough for the task, requiring on average \SI{1.33}{\kilo\byte} of storage per LiDAR scan against \SI{1.41}{\mega\byte} of raw~data.
To remedy such compact -- and thus feature-poor -- representation, we enhance state-of-the-art feature embedding and geometric-aware object matching \cite{wang2019dgcnn,qin2023geotransformer} with semantic information, both by encoding it in a network architecture, and by designing a semantically-informed loss to guide the training.
We further refine the estimates using \gls{ransac} and \gls{icp} on the object matches, obtaining average translational errors of about \SI{0.1}{\metre} and \SI{0.5}{\degree} respectively on KITTI \cite{behley2019semantickitti}.
Our system is outlined in \cref{fig:overview} and our principal contributions are:
\begin{enumerate}
\item An accurate registration approach working on extremely compact object-centric representations of LiDAR scans;
\item A semantic-enhanced neural network architecture and loss function for descriptorless object-matching;
\item Evaluation of the proposed methodology on KITTI \cite{geiger2012CVPR, behley2019semantickitti} and on a long-term localisation scenario between KITTI and KITTI-360 \cite{liao2022kitti}.
To the best of our knowledge, this is the first approach that considers long-term cross-dataset registration between these two.
\end{enumerate}

\section{Related Work}%
\label{sec:related}

Classical approaches to LiDAR scan registration rely on extracting descriptors from two point clouds, discovering point-to-point correspondences between them and exploiting them to regress the displacement.
One of the most widespread and simple approaches is \gls{icp} \cite{besl1992method}, which uses cartesian position to describe and match points.
%
As the environment changes through different viewpoints or over time, dense point-to-point correspondences can be replaced by more robust keypoint-to-keypoint correspondences.
As such, keypoint detectors have been developed to discover robust features in the environment and 3D descriptors to distinguish and match them.
Examples of classical, handcrafted keypoint detectors are ISS~\cite{zhong2009intrinsic} and KPG~\cite{mian2010repeatability}, which select salient points with large variations in their local neighbourhood.
Examples of 3D descriptors include \gls{fpfh}~\cite{rusu2009fpfh} and SHOT \cite{tombari2010unique}, which create histogram-like descriptors by taking into consideration the local topology of the neighbouring points.

\subsection{Learning-based Point Cloud Registration Losses}

Learning-based approaches have replaced classical ones, typically through end-to-end descriptors supervised by a differentiable \gls{svd} operation with ground-truth registration.
For instance, \gls{dcp} \cite{wang2019deep} adopts a point-based encoder to extract high-dimensional descriptors and a transformer-based head to compute soft matches to feed into an \gls{svd} module.
Similar to us -- but to subsample the point clouds to use in a \gls{dcp} setup -- DCPCR~\cite{wiesmann2022dcpcr} integrates a compression network, trained to downsample the points clouds while preserving the local information in the feature representation.

A different approach is to directly supervise the matching phase by cross-entropy or contrastive losses.
For instance, StickyPillars \cite{fischer2021stickypillars} inspired by the image-based SuperGlue \cite{sarlin2020superglue}, and MDGAT \cite{shi2021keypoint} build soft assignments supervised through ground-truth matches obtained by projection of the keypoints from one scan into the other.
Empirically, we show that a match-supervision approach tends to perform better in our specific case of few and feature-poor points.

\subsection{Learning-based Point Cloud Registration Architectures}

Architectures for point cloud processing for localisation adopt diverse learning architectures.
Indeed, whereas some works exploit the vast experience in image-based \glspl{cnn} \cite{chen2022overlapnet,choy2019fcgf}, most apply point- and, more recently, transformer- or graph-based approaches.

Although earlier approaches use purely point-based architectures \cite{aoki2019pointnetlk,pais20203dregnet,yuan2020deepgmr}, transformer-based architectures are recently enhancing them by applying self- and cross-attention in feature space to share context information from within or across LiDAR scans.
One example is GeoTransformer~\cite{qin2023geotransformer}, which uses KPConv-FPN \cite{thomas2019kpconv} and a custom transformer architecture with a keypoint descriptor invariant to rigid transformation.
Similarly, \gls{dcp} \cite{wang2019deep} and DCPCR~\cite{wiesmann2022dcpcr} exploit DGCNN \cite{wang2019dgcnn} and KPConv \cite{thomas2019kpconv} point~encoders respectively, and a transformer head for attention-based assignments.

Whereas transformers operate on fully-connected graphs, a purely graph-based method is, for instance, SEM-GAT~\cite{panagiotaki2023sem}, which exploits semantic and morphological features to find match candidates and compute match attention, leading to introspection capabilities \cite{panagiotaki2023semantic}.
Similarly, PREDATOR~\cite{huang2021predator} designs an overlap-attention module incorporating a \gls{gnn} connecting nearest-neighbour keypoints to extract and refine their feature descriptors.

In our proposed system, we apply a point encoder and transformer approach: we employ a DGCNN module for feature extraction, augmented by encoded semantic features together with the geometry of the scene, and a transformation-invariant matcher adapted from GeoTransformer \cite{qin2023geotransformer}.


\subsection{Use of Semantics in Point Cloud Registration}

Non-learned semantic approaches typically extract and describe objects and match them directly.
BoxGraph \cite{pramatarov2022boxgraph}, encodes objects with their bounding box achieving a very compact scan representation, while GOSMatch~\cite{zhu2020gosmatch} encodes histogram-based descriptors from the distances between the objects in the scene into vertex descriptors for initial pose estimation and verification.
For learned registration techniques we can distinguish three types of segmentation usage: augmenting geometric information with semantics, extracting object- or segment-level entities, and its usage in the loss term.
Examples of the first approach are DeepSIR~\cite{li2023deepsir} and SARNet~\cite{liu2022sarnet}, which learn both geometric and semantic features to constrain feature matching.
Differently, SegMatch~\cite{dube2017segmatch}, SegMap~\cite{segmap2019dube} and SemSegMap~\cite{cramariuc2021semsegmap} partition point clouds into higher-level segments and extract multidimensional descriptors for each segment, and match them across scenes.
More similar to us, InstaLoc~\cite{zhang2023instaloc} and SGPR~\cite{kong2020semantic} learn to segment and match individual objects to the prior scene.
In particular, \cite{kong2020semantic} applies a neural network architecture to extract vertex features based on graph connectivity.
Finally, methods like PADLoc~\cite{arce2023padloc} leverage panoptic segmentation of point clouds and use such information during training to aid the convergence to robust descriptors.
In our approach, we apply all three rules and extract object-level entities, which we couple with semantic class information as input \textit{and} as an additional term in the loss.
The resulting object features are orders of magnitude smaller than the object descriptors in the above-mentioned methods.

\section{Method}%
\label{sec:method}

\begin{figure*}[h]
\centering
\includegraphics[width=0.8\textwidth]{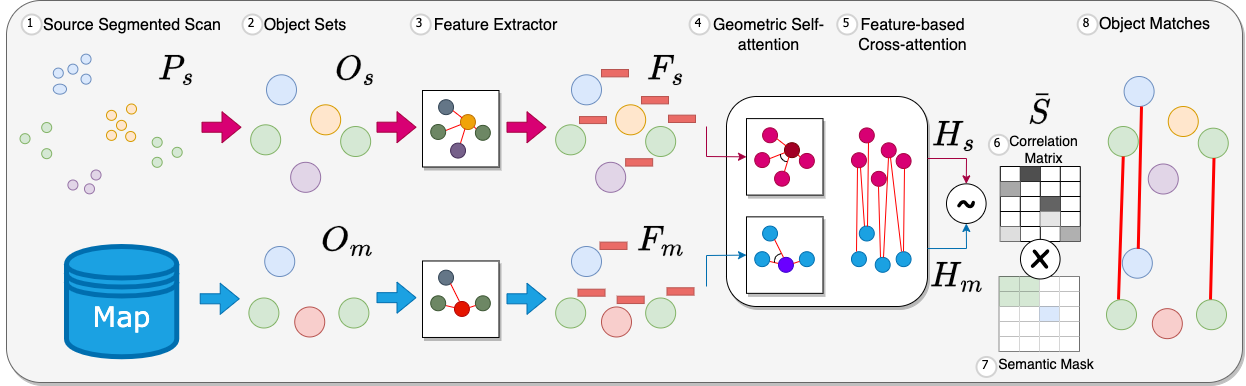}
\captionof{figure}
{
System diagram. Our method aims to register a semantically labelled query point cloud $P_s$ with a map.
It clusters $P_s$ into an object set $O_s$, keeping only instance centroids and semantic labels.
Then, the query and map sets are passed through a semantic embedding and feature extraction module.
The resulting object features $F_s$ and $F_m$ are then passed through a geometric self- and feature-based cross-attention matching module, producing a cross-correlation similarity matrix $\bar{S}$.
A semantic mask is then applied to filter erroneous matches, resulting in the final object correspondences.
\label{fig:method_diagram}}
\vspace{-.7cm}
\end{figure*}



We address the problem of robot pose estimation for large-scale outdoor LiDAR localisation with minimal, lightweight map and query representations.
We formulate this as a registration problem between two point clouds where we seek point correspondences to estimate the relative transformation.
By representing the point clouds as small sets of 3D object centroids and their respective semantic types, we obtain structures that yield extremely compact, highly-scalable semantic maps, yet allow very accurate localisation.
We employ a learned approach inspired by GeoTransformer~\cite{qin2023geotransformer}.
Still, rather than only relying on geometric features for superpoint matching, we integrate semantic information to obtain descriptive features for object matching as in~\cref{fig:method_diagram}.

\subsection{Problem Setup}

Consider a pair of point clouds, $P_s = \set{ p_i \given p_i \in \mathbb{R}^{3}}$ and $P_m = \set{p_j \given p_j \in \mathbb{R}^{3}}$, e.g., from a source live sensor stream of a vehicle and a pre-built target map.
We aim to estimate the relative transformation $\mathtt{T}_{s,m} = [\mathtt{R}_{s,m} | \mathbf{t}_{s,m}]$  where $\mathtt{R}_{s,m}\in SO(3)$ and $\mathbf{t}_{s,m}\in\mathbb{R}^3$ denote the rotation and translation.
We do so by generating key points, $K_s = \set{ k_i \given k_i \in \mathbb{R}^{3}}$ from $P_s$ and $K_m = \set{k_j \given k_j \in \mathbb{R}^{3}}$ from $P_m$.
Crucially, $K_s$ and $K_m$ are stable and visible across multiple revisits of an environment, and $|K_s| \ll |P_s|$ and $|K_m| \ll |P_m|$.
Given a set of ground-truth correspondences between the two sets of key points, $\mathtt{T}_{s,m}$ can be obtained using the Kabsch algorithm~\cite{kabsch1976svd} using \gls{svd}.
Hence, we aim to obtain an optimal set of correspondences $\mathcal{C} = \set{ (k_i, k_j) \given k_i \in K_s, k_j \in K_m }$ to produce the final pose estimate.
We show that using semantic object-like instances as key points provides a consistent basis for pose estimation even across long-term revisits of the same place under variation in object layout.

\subsection{Object Extraction}

We follow BoxGraph~\cite{pramatarov2022boxgraph} and SGPR~\cite{kong2020semantic}, where input point clouds are segmented via a pre-trained semantic segmentation network and then clustered into object instances based on their semantic labels and Euclidean coordinates.
Panoptic segmentation or object detection might yield more accurate instances, yet we opt for this simpler approach since datasets with static object labels are scarce.

In particular, consider a point cloud $P$ (\ding{172}~in~\cref{fig:method_diagram}) and the set $L$ containing a semantic label for each point in $P$; here, $L = \set{l_i \given l_i\in\mathbb{L} }$, where $\mathbb{L}\subset\mathbb{N}$ is a finite set of semantic classes.
We apply the \gls{dbscan} algorithm to extract a set of object-like clusters.
For each cluster we keep its centroid $o$ -- as the mean coordinate of its points -- and semantic label $l$.
We then end up with with a labelled object set $O = \set{(\mathbf{o}_i, l_i) \given \mathbf{o}_i\in\mathbb{R}^3, l_i\in\mathbb{L} }$, which we use as keypoints (\ding{173}~in~\cref{fig:method_diagram}).
Contrary to other approaches that extract object features by exploiting a point cloud's intra- and inter-object geometric structure, we instead utilise only the semantic label and the inter-object relationships, and \textit{learn} any further representations associated with the object.

\subsection{Feature Enhancement}

Indeed, we encode each object's semantic class and neighbouring geometric structure to extract discriminative object descriptors (\ding{174}~in~\cref{fig:method_diagram}).
For this, we apply a learnable function $f_{sem} = h_{sem} \circ e_{emb}$ to each cluster's semantic label $l_i$, where $e_{emb}: \mathbb{L} \rightarrow \mathbb{R}^{d_{emb}}$ is a categorical embedding function and $h_{sem}: \mathbb{R}^{d_{emb}} \rightarrow \mathbb{R}^{d_{sem}}$ is parametrised by a small MLP.
We then enhance the features with structural context $\mathbf{o}_i$ through a three-layer DGCNN-based module $E$~\cite{wang2019dgcnn}, producing $F = \set{f_i \given f_i = E(\mathbf{o}_i \Vert f_{sem}(l_i)), (\mathbf{o}_i, l_i) \in O, f_i n \in \mathbb{R}^{d_f}}$, where $(\cdot \Vert \cdot)$ denotes concatenation.
At each layer, the feature $x_i$ of the $i$-th point gets transformed into the feature $x'_i$ through a projection network $h_{ec}$ and max-pooling layer as in:
\begin{equation}
\begin{split}
x'_i = \max_{j \in \mathrm{kNN}(i)} h_{ec}(x_i \mathbin\Vert x_j - x_i)
\end{split}
\end{equation}
where $\mathrm{kNN}(i)$ are $i$'s $k$ neighbours in feature space. 
Whilst $x_i$ and $(x_j - x_i)$ are not transformation invariant yet, the operation informs the resulting features of the local structure.


\subsection{Object Similarity and Matching}

The feature vector of each object must be sufficiently descriptive and discriminative to be effectively matched across scans.
Since we discard the internal structure of each object instance, it is crucial to exploit the scene's structure by incorporating it into each object's features.
Moreover, as the localisation setting is agnostic of the relative transformation between the scans, so must be the object descriptors.

Since cross-correlation between features across scenes benefits the matching task~\cite{wang2019deep, sarlin2020superglue, qin2023geotransformer}, we employ the Superpoint Matching Module of GeoTransformer~\cite{qin2023geotransformer}, which explicitly models the layout within scenes through a geometric self-attention module (\ding{175}~in~\cref{fig:method_diagram}), and the latent feature similarity across scenes through a feature-based cross-attention module (\ding{176}~in~\cref{fig:method_diagram}).
The geometric self-attention module encodes the global structure of the scene in a transformation-invariant manner by operating on the relative distances between pairs and the relative angles between triplets of objects.
The feature-based cross-attention module, instead, facilitates the feature exchange between the two scenes.
Further details can be found in the original paper~\cite{qin2023geotransformer}.
The self- and cross-attention modules are interleaved $N_t$ times, and, from $O_s$, $O_m$ and the corresponding object features above, output the hybrid features $H_s = \set{h^s_i \given h^s_i \in \mathbb{R}^{d_h}}$ and $H_m = \set{h^m_j \given h^m_j \in \mathbb{R}^{d_h}}$, suitable for matching.
Following \cite{qin2023geotransformer}, they produce a normalised Gaussian correlation matrix $\bar{S} \in \mathbb{R}^{|O_s| \times |O_m|}$ which serves as matching scores (\ding{177}~in~\cref{fig:method_diagram}). 

At this point, we extend \cite{qin2023geotransformer} and exploit the semantics of the scene to refine this score further by filtering mismatched objects.
We mask out the similarity scores of objects with different labels (\ding{178}~in~\cref{fig:method_diagram}), i.e. set $\bar{s}_{i,j} \coloneqq 0$, if $l^s_i \neq l^m_j$.
The optimal set of object correspondences is then obtained by selecting the top $N_c$ matches with the highest similarity:
\begin{equation}
\mathcal{\hat{C}} = \set{ (\mathbf{o}_i, \mathbf{o}_j) \given \mathbf{o}_i \in O_s, \mathbf{o}_j \in O_m, (i, j) \in \mathrm{topk}_{i,j}(\bar{S}) }
\end{equation}









\subsection{Pose Estimation and Refinement}

To find the relative transformation $\mathtt{T}_{s,m} = [\mathtt{R}_{s,m} | \mathbf{t}_{s,m}]$ between object sets $O_s$ and $O_m$ given the above correspondences (\ding{179}~in~\cref{fig:method_diagram}), we solve
\begin{equation}
\begin{split}
\hat{\mathtt{R}}_{s,m}, \hat{\mathbf{t}}_{s,m} = \min_{\mathtt{R}, \mathbf{t}} \sum_{(\mathbf{o}_i, \mathbf{o}_j) \in \mathcal{\hat{C}}} \bar{S}_{i, j} \norm{\mathtt{R} \mathbf{o}_i + \mathbf{t} - \mathbf{o}_j}^2_2
\end{split}
\end{equation}
directly via the weighted SVD algorithm.
Alternatively, we can apply the \gls{ransac} algorithm that iteratively selects a subset of matches and tries to maximise the number of inliers up to a distance tolerance for a fixed set of iterations.
We explore both approaches in the experimental evaluation; yet, it is important to note that, whereas \gls{ransac} is computationally expensive for dense point clouds, its added complexity is negligible given the low cardinality of the object sets $O_s$ and $O_m$.
Finally, we find it helpful to refine further the relative transformation with~\gls{icp}, reducing the noise in the coarse estimate above, which again adds insignificant computational overhead.

\subsection{Loss Function}

Deep registration approaches are typically supervised in two different ways: a regression loss on the final pose estimate that exploits the differentiability of the soft-assignment matrix between source and target point clouds~\cite{wang2019deep,wiesmann2022dcpcr,arce2023padloc}, or a cross-entropy/contrastive loss which uses ground-truth matches to directly supervise the similarity scores~\cite{sarlin2020superglue, fischer2021stickypillars}.
Empirically, when key points are sparse, as in our case, even small ambiguity in the soft assignment matrix introduces large errors in the final estimate; thus, we opt to supervise the object features in a metric fashion similar to GeoTransformer~\cite{qin2023geotransformer, sun2020circle}, which employs an overlap-aware circle loss on the superpoint features.

The overlap-aware circle loss aims to bring together the features of positive pairs of objects, i.e. spatially proximal, and push apart those of negative pairs, i.e., objects far in space.
It does so by weighting the positive matches according to the overlap ratio of their clusters.
As we discard object geometries, we cannot recover a measure of overlap.
For this reason, we introduce a semantic distance-aware circle loss which focuses the learning on spatially-proximal object pairs proportionally to their Euclidean distance, \textit{and} ensures the positive pairs are of objects of the same semantic class.

Formally, given an anchor object $\mathbf{o}_i \in O_s$, we consider an object $\mathbf{o}_j \in O_m$ a positive if $d_{i,j} \coloneqq \norm{\mathbf{o}_i - \mathbf{o}_j}^2_2 < \tau_{match}$ and $l_i == l_j$.
We set $\tau_{match} \coloneqq \SI{1}{\metre}$ and denote the set of anchor objects with $\mathcal{A}_s \subset O_s$, the set of positive corresponding objects with $\mathrm{pos}(i)  \subset O_m$, and the rest as negatives with $\mathrm{neg}(i) \subset O_m$.
If we let the distance ratio be $\rho_{i,j} \coloneqq 1 - \frac{d_{i,j}}{\tau_{match}} \in [0, 1]$ for positive matches, we can formulate the semantic distance-aware circle loss with respect to $\mathcal{A}_s$ as:
\begin{equation}
\vspace{-5pt}
\begin{multlined}
\mathcal{L}^s_{DC} = \frac{1}{|\mathcal{A}_s|} \sum_{i \in \mathcal{A}_s} \log \left( 1+ \sum_{j \in \mathrm{pos}(i)} e^{\beta^p_{i,j} ( h_{i,j} - \Delta_p )} \right. \cdot \\
\left. \sum_{j \in \mathrm{neg}(i)} e^{\beta^n_{i,k} ( \Delta_n - h_{i,k} )} \right)
\end{multlined}
\end{equation}
where $h_{i,j} = \norm{h^s_i - h^m_j}^2_2$ is the distance in feature space. The positive pairs and negative pairs are weighted according to weight factors $\beta^p_{i,j} = \sqrt{\rho_{i,j}} \gamma( h_{i,j} - \Delta_p )$ and $\beta^n_{i,k} = \gamma( \Delta_n - h_{i,k} )$, where $\gamma = 40$ is a scaling factor and $\Delta_p = 0.1$ and $\Delta_n = 1.4$ are the corresponding margins.
We define the loss $\mathcal{L}^m_{DC}$ for the target objects $\mathcal{A}_m$ analogously, yielding the overall loss $\mathcal{L}_{DC}$ as the average of $\mathcal{L}^s_{DC}$ and $\mathcal{L}^m_{DC}$.

\section{Experimental Setup}%
\label{sec:experiments}

\subsection{Datasets}

We evaluate our method on KITTI~\cite{behley2019semantickitti} and KITTI-360 \cite{liao2022kitti}, which include traversal sequences of different areas in Karlsruhe, Germany, by a vehicle equipped with a Velodyne HDL-64 LiDAR sensor.
We first explore short-term revisits -- applicable to loop closing and SLAM scenarios -- on KITTI sequence \texttt{08}, challenging due to reverse revisits.
The second considers long-term revisits in a teach\&repeat setup where we select KITTI sequence \texttt{07} as a map and localise KITTI-360 sequence \texttt{09}, which has been recorded two years apart, with potentially significant changes in the scenes.
In all our evaluations, we input to our model semantic labels as predicted by a top-performing segmentation network, Cylinder3D~\cite{zhou2020cylinder3d}, pre-trained on SemanticKITTI~\cite{behley2019semantickitti}.



We consider the KITTI sequences that contain revisits and use \texttt{00}, \texttt{05}, \texttt{06}, and \texttt{09} for training and \texttt{02} for validation, by selecting pairs of scans within \SI{3}{\metre} that are at least 50 frames apart, as in~\cite{kong2020semantic,li2021ssc,pramatarov2022boxgraph}. 
While \SI{3}{\metre} may be a limited threshold for general registration, we argue that a coarse location estimate could be achieved with raw GPS readings or place recognition approaches such as SGPR~\cite{kong2020semantic}, which use a similar compact object representation as ours.

\begin{figure}[]
\centering
\includegraphics[width=0.5\columnwidth]{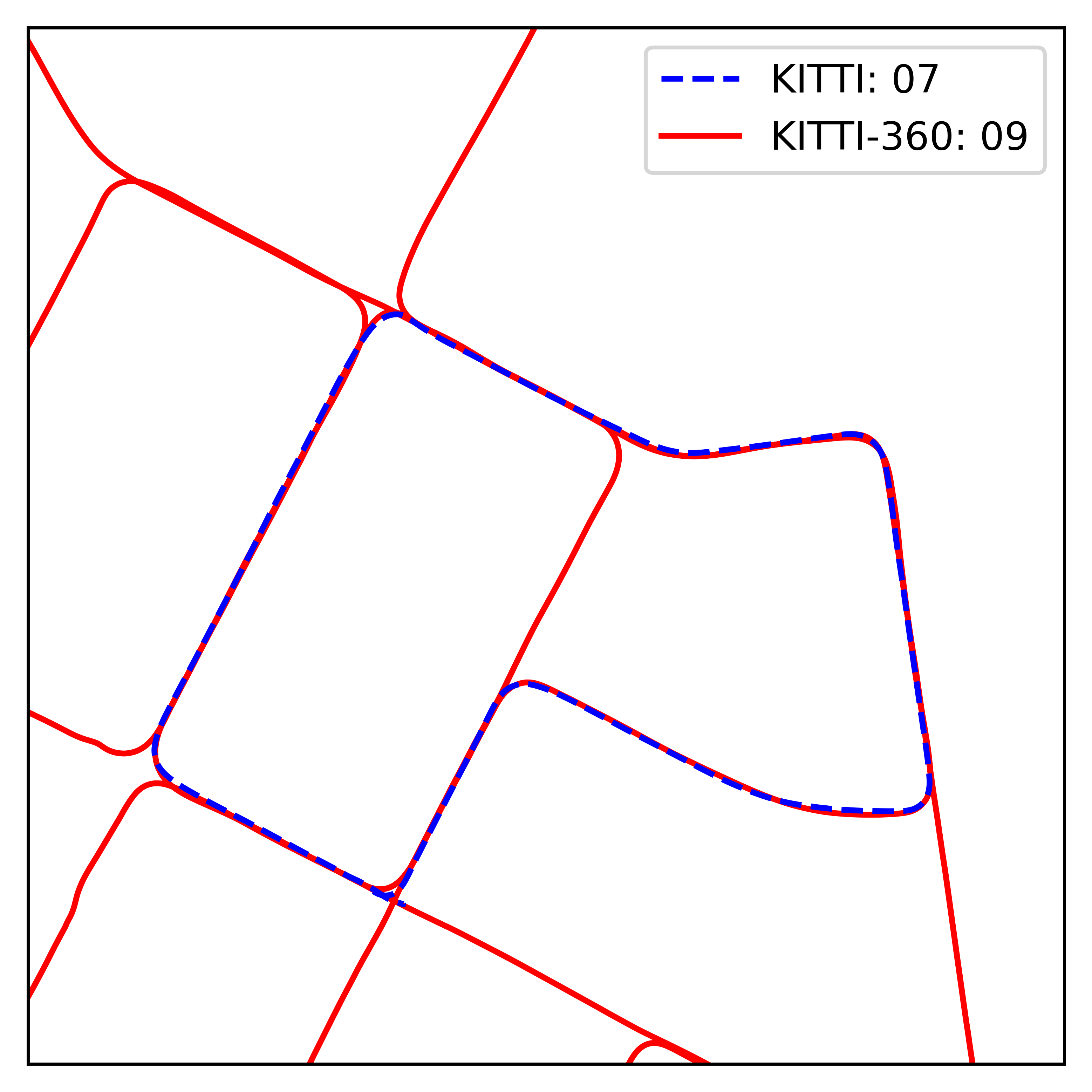}
\caption{
KITTI-360 sequence \texttt{09}, registered on KITTI sequence \texttt{07}.
\label{fig:localisation_experiment}}
\vspace{-.75cm}
\end{figure}

The raw ground-truth poses are noisy, so we further refine them via~\gls{icp}.
To register KITTI and KITTI-360, we use the raw GPS measurements to select pairs within~\SI{3}{\metre}.
Then we remove points belonging to dynamic objects and the road class using Cylinder3D predictions, and again refine with~\gls{icp} to obtain accurate ground-truth poses, as in \cref{fig:localisation_experiment}.

\subsection{Baselines}

We compare our method with geometric, learning-based and semantic approaches.
\begin{enumerate*}
    \item \gls{ransac}-based, handcrafted~\gls{fpfh} features as a dense geometric approach -- with downsampled input point clouds with a voxel size of \SI{30}{\centi\metre};
    \item BoxGraph~\cite{pramatarov2022boxgraph}, a compact hand-crafted semantic method;
    \item PADLoC~\cite{arce2023padloc}\footnote{PADLoC's public weights are trained on a slightly larger superset of our training data, containing pairs within \SI{4}{\metre}, and including sequence \texttt{07}.}, a learned semantic global localisation approach;
    \item superpoint-matching module of the vanilla GeoTransformer~\cite{qin2023geotransformer} (here, GeoTF-SP). We re-train the original architecture on our training data and, at test time, replace the fine registration module on the dense point patches with RANSAC-based matching on superpoint correspondences.
\end{enumerate*}
We also refine all methods with \gls{icp} for a fair comparison.

\subsection{Metrics}

To measure the performance, we report the \gls{rte} and \gls{rre}:
\begin{equation}
    \mathrm{RTE} = \norm{\hat{\mathbf{t}} - \mathbf{t}}_2, 
    \mathrm{RRE} = \cos^{-1}\left(\frac{\text{Tr}\:(\hat{\mathtt{R}}^T \mathtt{R}) - 1}{2} \right)
\end{equation}
between the estimated $[\hat{\mathtt{R}}_{s,m}|\hat{\mathbf{t}}_{s,m}]$ and ground-truth $[\mathtt{R}_{s,m}| \mathbf{t}_{s,m}]$ rotation and translation, averaged over successful registrations.
Subscript is removed for brevity.
We also report the registration recall that measures the percentage of successful registrations such that $\mathrm{RTE} < \tau_t$ and $\mathrm{RRE} < \tau_R$. 

\subsection{Implementation Details}

As in~\cite{pramatarov2022boxgraph}, we only keep static objects (sidewalk, building, fence, vegetation, trunk, pole, and traffic sign), stable across revisits.
Our semantic embedding dimension is $4$, followed by an MLP(32, 64, 128).
We use 3 EdgeConv(64, 64) layers in DGCNN with 30 nearest neighbours, followed by an MLP(1024, 512, 256, 256) with ReLU.
Normalisation and dropout deteriorate performance, so we omit them.
GeoTransformer's matching module is set as in~\cite{qin2023geotransformer}, with 3 sets of self- and cross-attention layers with 4 attention heads, and output dimension of $256$.
We train on an NVIDIA RTX 3090 Ti GPU with a batch size of $32$ for $50$ epochs, a learning rate of $1\cdot10^{-3}$ and ADAM optimizer, halving the learning rate whenever the loss has plateaued for 5 epochs.
We apply random yaw rotations up to $360^\circ$, randomly subsample raw point clouds to $24000$ points before clustering and add random jitter to object centroids to simulate sensor noise.
At inference time, we set the number of object correspondences $N_c$ to $15$ when using~\gls{svd}, and $60$ with~\gls{ransac}.
This allows~\gls{svd} to focus on high-quality matches, while~\gls{ransac} has higher outlier tolerance by design.
Segmentation and clustering take \SI{62}{\milli\second} and \SI{80}{\milli\second}, respectively, while registering a pair of scans takes \SI{20}{\milli\second} including refinement, showing the efficiency of the approach.

\section{Results}

\subsection{Map Compactness}

Here we compare our storage requirements to other methods.
Our representations are extremely compact, producing on average 105 and maximum 238 objects-like instances per point cloud, similar to BoxGraph~\cite{pramatarov2022boxgraph}.
For each instance, we store the three floating point coordinates of the centroid and a byte for the semantic class, which on average requires \SI{1.33}{\kilo\byte} of storage.
With our representation, the \SI{3.2}{\kilo\metre}-long KITTI sequence~\texttt{08} can be stored with~\SI{4.47}{\mega\byte} only.

BoxGraph stores three additional floating point numbers -- the objects' bounding boxes -- almost doubling the memory requirement.
The difference is even larger compared to dense methods like GeoTranformer \cite{qin2023geotransformer}, which require the full LiDAR scan: given 120k as the number of sampled points, common in \gls{ad} applications, these methods would require three floats for point, i.e. \SI{1.4}{\mega\byte}.
Even when voxelised representations are used to reduce the point count -- reaching usually 20k points -- storage requirements are about \SI{234.4}{\kilo\byte}.
Compressed methods require an intermediate storage space: for instance, OctSqueeze \cite{huang2020octsqueeze} ranges between 2 and 15 bits per point (bpp), thus requiring from \SI{29.3}{\kilo\byte} to \SI{219.7}{\kilo\byte}.
DCPCR, instead, can compress a point cloud 100-folds\footnote{As DCPCR uses aggregated point clouds; here we use the ratio they report for comparison.}, compared to our average ratio of about 1:1000.


\subsection{Pose Estimation}
\label{sec:exp:poseest}

We first explore in \cref{tab:registration} the pose estimation performance of our approach in the short-term setting, where revisits happen within the same sequence.
Both our method and BoxGraph take semantic labels as inputs explicitly, so entries with (GT) represent results with ground-truth \textit{semantic} labels from SemanticKITTI, as opposed to Cylinder3D predictions.

Whilst classical RANSAC-based approaches struggle to deduce accurate pose, we achieve accurate results with RANSAC and SVD.
In particular, our method using Cylinder3D semantic labels estimates the pose within \SI{0.3}{\metre} and $1^\circ$ more than $85\%$ of the time.
BoxGraph struggles with reverse revisits when using Cylinder3D labels, even after ICP-refinement.
Its performance improves significantly when using ground-truth semantics, so we conclude it strongly depends on the quality of input segmentation.
In contrast, our method shows greater stability when using ground-truth semantic labels as input.
In addition, using the vanilla GeoTF-SP (with RANSAC) proves ineffective in this setting, even though its deeper features are directly supervised for matching.
This justifies our use of object centroids as distinctive and repeatable keypoints.

We observe similar trends in the second setting, where we investigate the long-term localisation performance.
We see that PADLoC, which incorporates semantics implicitly, struggles to register reliably without~\gls{icp} refinement.
This might mean it is difficult to generalize the semantic features across long periods.
Our method models objects explicitly and demonstrates high recall of pose estimates with errors within \SI{0.5}{\metre} and $5^\circ$.
Note that neither Cylinder3D, nor our method are trained on KITTI-360, showing further robustness across large temporal intervals.
This is in line with the other explicit object-based method, BoxGraph, whose performance improves.
Comparing the different variants of our method, we see that weighted~\gls{svd} achieves lower metric errors since it operates on high-quality matches, while~\gls{ransac} indeed maximises the consensus between matches, yielding high recall.

\begin{table}[h]
\centering
\vspace{-5pt}
\resizebox{0.95\columnwidth}{!}{
\begin{tabular}{r|l|rrr|rrr}

\toprule
 & & \multicolumn{3}{c|}{\textbf{\SI{0.3}{\metre}/\SI{1}{\degree}}} & \multicolumn{3}{c}{\textbf{\SI{0.5}{\metre}/\SI{5}{\degree}}} \\
 & \textbf{Method} & \textbf{RR} & \textbf{RTE} & \textbf{RRE} & \textbf{RR} & \textbf{RTE} & \textbf{RRE} \\
\midrule

\multirow{14}{*}{\rotatebox[origin=c]{90}{Seq: \texttt{08}}}
 & FPFH & 3.48 & 0.18 & 0.63 & 15.42 & 0.28 & 1.59 \\
 & GeoTF-SP & 20.29 & 0.19 & 0.52 & 42.26 & 0.27 & 0.79 \\
 & GeoTF-SP + ICP & 20.49 & 0.19 & 0.51 & 40.83 & 0.26 & 0.79 \\
 & PADLoC & 14.24 & 0.20 & 0.66 & 67.78 & 0.30 & 1.21 \\
 & PADLoC + ICP & 49.28 & \textbf{0.11} & \textbf{0.28} & 80.02 & 0.20 & 0.56 \\
 & BoxGraph & 18.75 & 0.16 & 0.60 & 44.67 & 0.22 & 1.16 \\
 & BoxGraph + ICP & 53.28 & 0.12 & 0.34 & 59.38 & 0.13 & 0.45 \\
 & Ours - RANSAC & \textbf{87.50} & 0.13 & 0.45 & \textbf{99.80} & 0.14 & 0.54 \\
 & Ours - SVD & 85.66 & 0.12 & 0.44 & 95.18 & \textbf{0.12} & \textbf{0.52} \\[1pt]
 \cline{2-8}
 &\\[-5pt]
 & BoxGraph (GT) & 51.23 & 0.12 & 0.50 & 77.61 & 0.16 & 0.85 \\
 & BoxGraph + ICP (GT) & 80.64 & 0.11 & 0.40 & 85.09 & 0.11 & 0.44 \\
 & Ours - RANSAC (GT) & 88.52 & 0.11 & 0.45 & \textbf{99.49} & 0.13 & 0.52 \\
 & Ours - SVD (GT) & \textbf{89.60} & \textbf{0.09} & \textbf{0.41} & 96.52 & \textbf{0.10} & \textbf{0.47} \\

\midrule
\midrule

\multirow{9}{*}{\rotatebox[origin=c]{90}{Seq: \texttt{09}, Map: \texttt{07}}}
 & FPFH & 42.47 & 0.16 & 0.48 & 72.34 & 0.22 & 0.93 \\
 & GeoTF-SP & 38.22 & 0.17 & 0.38 & 56.01 & 0.23 & 0.51 \\
 & GeoTF-SP + ICP & 39.32 & 0.13 & 0.31 & 54.00 & 0.20 & 0.45 \\
 & PADLoC & 0.01 & 0.23 & 0.84 & 5.22 & 0.38 & 3.31 \\
 & PADLoC + ICP & 46.93 & 0.13 & \textbf{0.30} & 87.00 & 0.21 & 0.73 \\
 & BoxGraph & 41.92 & 0.14 & 0.44 & 60.20 & 0.19 & 0.74 \\
 & BoxGraph + ICP & 62.52 & 0.12 & 0.32 & 71.29 & 0.15 & 0.41 \\
 & Ours - RANSAC & 78.14 & \textbf{0.11} & 0.35 & \textbf{94.14} & 0.15 & 0.47 \\
 & Ours - SVD & \textbf{80.52} & \textbf{0.11} & \textbf{0.30} & 92.66 & \textbf{0.14} & \textbf{0.39} \\

\bottomrule
\bottomrule
\end{tabular}
}
\caption{Registration errors $[\si{\metre}|\si{\degree}]$ at different thresholds for short-term revisits on KITTI sequence \texttt{08} and long-term localisation on KITTI-360 sequence \texttt{09} with KITTI sequence \texttt{07} as map.   \label{tab:registration}}
\vspace{-.5cm}
\end{table}

\begin{table}[t]
\footnotesize
\centering
\renewcommand{\arraystretch}{1.2}
\resizebox{\columnwidth}{!}{\begin{tabular}{lcccccccc}
    \toprule
    & & \multicolumn{3}{c}{\textbf{Architecture}} & \multicolumn{3}{c}{\textbf{Metric Errors [\SI{0.3}{\metre}/\SI{1}{\degree}]}} \\
     & \textbf{Sem. Emb.} & \textbf{DGCNN} & \textbf{Matching Module} & \textbf{Loss} & \textbf{RR} & \textbf{RTE} & \textbf{RRE} \\
    \midrule
    (1) & \xmark & \checkmark & GeoTF & $\mathcal{L}^{-}_{DC}$ & 36.17 & 0.14 & 0.40 \\ 
    (2) & \xmark & \checkmark & GeoTF + ICP & $\mathcal{L}^{-}_{DC}$ & 68.75 & \textbf{0.11} & \textbf{0.38} \\ 
    (3) & \xmark & \checkmark & GeoTF & $\mathcal{L}_{DC}$ & 26.59 & 0.15 & 0.46 \\ 
    (4) & \checkmark & \checkmark & GeoTF & $\mathcal{L}_{SVD}$ & 0.31 & 0.21 & 0.70 \\ 
    (5) & \checkmark & \checkmark & GeoTF & $\mathcal{L}^{-}_{DC}$ & 31.25 & 0.14 & 0.40 \\ 
    (6) & \checkmark & \xmark & GeoTF & $\mathcal{L}_{DC}$ & 17.01 & 0.17 & 0.56 \\ 
    (7) & \checkmark & \checkmark & KPConv + TF & $\mathcal{L}_{SVD}$ & 16.09 & 0.13 & 0.57 \\ 
    (8) & \checkmark & \checkmark & GeoTF & $\mathcal{L}_{DC}$ & 41.09 & 0.14 & 0.42 \\ 
    (9) & \checkmark & \checkmark & GeoTF + ICP & $\mathcal{L}_{DC}$ & \textbf{75.67} & \textbf{0.11} & 0.41 \\ 
    \bottomrule
    \bottomrule
\end{tabular}}
\caption{Ablation on architecture on KITTI sequence \texttt{08}, $[\si{\metre}|\si{\degree}]$. \label{tab:ablation_architecture}}
\vspace{-17pt}
\end{table}



\subsection{Ablation Studies and Analysis}
\label{sec:exp:ablation}

We analyse the different components of our method to gain further insight into our design choices.
We report results on KITTI sequence \texttt{08} and use weighted SVD on $N_c = 60$ matches.
\gls{svd} is more strongly affected by erroneous matches, so this better models the match quality.

\Cref{tab:ablation_architecture} reports the results, where $\mathcal{L}_{DC}$ denotes the proposed distance-aware circle loss and $\mathcal{L}^{-}_{DC}$ ignores semantic classes and allows mislabelled matches, while $\mathcal{L}_{SVD} = ||\mathbf{t}_{gt} - \hat{\mathbf{t}}|| + ||\mathbf{I} - \mathbf{R}^\mathbf{t}_{gt}\hat{\mathbf{R}}||$ denotes the explicit pose loss estimated with SVD on the full similarity matrix $\bar{S}$ (\ding{177}~in~\cref{fig:method_diagram}).

We can see in (1)-(5) that performance deteriorates without the input semantic embeddings and the semantic filtering in our loss.
We notice that having no input semantics and filtering in the loss, (1)-(2), performs better than omitting either one of them (3)-(5), potentially meaning that having semantics at only one end confuses the network while ignoring them altogether allows it to learn matches purely based on spatial proximity.
The largest drop in performance happens when replacing the $\mathcal{L}_{DC}$-loss (4), which justifies our design choice.
(6) also shows that the DGCNN feature enhancement is crucial for learning discriminative object features.
In (7) we replace the GeoTransformer Superpoint Matching Module with a KPConv-based feature encoder and a feature-based transformer head as in~\cite{wiesmann2022dcpcr}.
The deteriorated performance shows the descriptive power of the geometric self-attention of GeoTransformer, which supports our design.

\subsection{Reducing Registration Overlap}
\label{sec:exp:increased_range}

So far we consider pairs of point clouds with sufficient overlap.
We now briefly study the performance of our method with lesser overlap.
We follow a common setup~\cite{qin2023geotransformer,choy2019fcgf,huang2021predator} which aims to register point cloud pairs that are at least \SI{10}{\metre} apart.
We train on KITTI sequences \texttt{00}-\texttt{05} and report the averaged results on sequences~\texttt{08}-\texttt{10} in \cref{tab:pairs_10}.
We see that while our performance is limited due to the low quality of object instances we extract, our RANSAC-based approach still obtains more than $90\%$ registration recall within the commonly used boundaries of $\tau_t = \SI{2}{\metre}$ and $\tau_R = 5^\circ$.
This is a significant improvement compared to BoxGraph which uses the same instances as our approach, and showcases the potential of learned centroid-only object-based registration.

\begin{table}[h]
\centering
\vspace{-5pt}
\begin{tabular}{r|rrr}

\toprule
 & \multicolumn{3}{c}{\SI{2}{\metre}/\SI{5}{\degree}} \\
 \textbf{Method} & \textbf{RR} & \textbf{RTE} & \textbf{RRE} \\
\midrule
BoxGraph & 10.85 & 0.61 & 1.95 \\
 Ours - RANSAC & \textbf{95.53} & \textbf{0.27} & \textbf{1.05} \\
 Ours - SVD & 67.97 & 0.33 & 1.07 \\
\midrule BoxGraph (GT) & 16.49 & 0.52 & 1.84 \\ Ours - RANSAC (GT) & \textbf{92.65} & \textbf{0.26} & \textbf{1.06} \\
 Ours - SVD (GT) & 65.90 & 0.27 & 0.96 \\

\bottomrule
\bottomrule
\end{tabular}
\caption{Registration errors $[\si{\metre}|\si{\degree}]$ for scans at least \SI{10}{\metre} apart, averaged over KITTI sequences \texttt{08}-\texttt{10}.
\label{tab:pairs_10}}
\vspace{-.5cm}
\end{table}


\section{Conclusion}%
\label{sec:conclusion}

We presented a novel approach for global registration of LiDAR point clouds with extremely compact object representations.
We show that using only the 3D centroids and semantic type of object instances is enough to estimate accurate poses in challenging conditions, including reverse and long-term revisits of the same place.
We employ a low-level geometric-attention-based matching module, and enhance it with high-level semantics to increase its discriminative power.
The resulting maps have very low storage requirements, which enables drastic scaling of the mapped areas.
In addition, working directly on objects can be a subject for cross-modal localisation or semantic analysis and reasoning, which is the focus of future work.

\bibliographystyle{IEEEtran}
\bibliography{biblio}

\end{document}